\documentclass[10pt,twocolumn,letterpaper]{article}

\usepackage{wacv}
\usepackage{times}
\usepackage{epsfig}
\usepackage{graphicx}
\usepackage{amsmath}
\usepackage{amssymb}
\usepackage{float}      % H float option
\usepackage{esvect}     % Vectors
\usepackage{gensymb}    % Degrees
\usepackage{bm}         % Bold math
\usepackage{tabularx}   % Fancy tables
\usepackage{caption}    % Finer control over captions/labels

\newcommand{\predpose}{\hat{\vv{p}}}       
\newcommand{\gtpose}{\vv{p}}      
\newcommand{\predpos}{\hat{\vv{x}}}       
\newcommand{\gtpos}{\vv{x}}         
\newcommand{\predrot}{\hat{\vv{q}}}       
\newcommand{\gtrot}{\vv{q}}    
\newcommand{\deltavec}{\vv{d}}  

\newcolumntype{P}[1]{>{\centering\arraybackslash}p{#1}}     % Command for centering columns of given width

\usepackage[pagebackref=true,breaklinks=true,letterpaper=true,colorlinks,bookmarks=false]{hyperref}

\wacvfinalcopy % *** Uncomment this line for the final submission

\def\wacvPaperID{755} % *** Enter the wacv Paper ID here

\ifwacvfinal\fi
\begin{document}

%%%%%%%%% TITLE
\title{Improving Image-Based Localization with Deep Learning:\\The Impact of the Loss Function}

% Authors at the same institution
\author{Isaac Ronald Ward \hspace{2cm} M. A. Asim K. Jalwana \hspace{2cm} Mohammed Bennamoun \\
University of Western Australia, Perth, Australia\\
{\tt\small isaac.ward@uwa.edu.au}
}

\maketitle
\ifwacvfinal\thispagestyle{empty}\fi

% Dataset available here 
% https://github.com/anon-datasets/gemini

%%%%%%%%% ABSTRACT
%Leave two blank lines after the Abstract, then begin the main text.
\begin{abstract}
%Talk more a bout loss stuff
This work investigates the impact of the loss function on the performance of Neural Networks, in the context of a monocular, RGB-only, image localization task. A common technique used when regressing a camera's pose from an image is to formulate the loss as a linear combination of positional and rotational mean squared error (using tuned hyperparameters as coefficients). In this work we observe that changes to rotation and position mutually affect the captured image, and in order to improve performance, a pose regression network's loss function should include a term which combines the error of both of these coupled quantities. Based on task specific observations and experimental tuning, we present said loss term, and create a new model by appending this loss term to the loss function of the pre-existing pose regression network `PoseNet'. We achieve improvements in the localization accuracy of the network for indoor scenes; with decreases of up to $26.7$\% and $24.0$\% in the median positional and rotational error respectively, when compared to the default PoseNet.
\end{abstract}

%%%%%%%%% INTRODUCTION
\section{Introduction}

In Convolutional Neural Networks (CNNs) and other Neural Network (NN) based architectures, a `loss' function is provided which quantifies the error between the ground truths and each of the NN's predictions. This scalar quantity is used during the backpropagation process, essentially `informing' the NN on how to adjust its trainable parameters. Naturally, the design of this loss function greatly affects the training process, yet simple metrics such as mean squared error (MSE) are often used in place of more intuitive, task specific loss functions. In this work, we explore the design and subsequent impact of a NN's loss function in the context of a monocular, RGB-only, image localization task.

\begin{figure}[H]
    \begin{center}
        \includegraphics[width=0.95\linewidth]{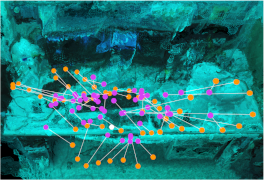}
    \end{center}
    
    \caption{A sample of the predicted pose positions (purple) generated for the ground truth poses (orange) in the \textit{7Scenes} \textit{Heads} scene using our proposed model. The scene's origin (white) and SfM reconstruction is rendered for reference. Image best viewed in color. }
    \label{fig:chessresults}
\end{figure}

The problem of image localization --- that is; extracting the position and rotation (herein referred to collectively as the `pose') of a camera, directly from an image --- has been approached using a variety of traditional and deep learning based techniques in the recent years.  The problem remains exceedingly relevant as it lies at the heart of numerous technologies in Computer Vision (CV) and robotics, \eg geo-tagging, augmented reality and robotic navigation.

More colloquially, the problem can be understood as trying to find out where you are, and where you are looking, by considering only the information present in an RGB image. 

%In the literature this is sometimes referred to as \textbf{the kidnapped robot problem}: ``how can a device re-localize itself with any certainty from an arbitrary, unknown location''. Since a camera is capturing the input image, the terms image localization and camera (re)localization are often used interchangeably. Pipelines which localize from image inputs only can also be referred to as Visual Positioning Systems (VPS).

CNN based approaches to image localization --- such as PoseNet \cite{kendall_posenet_2015} --- have found success in the recent years due to the availability of large datasets and powerful training hardware, but the performance gap between these systems and the more accurate SIFT feature-based pipelines remains large. For example, the SIFT-based Active Search algorithm \cite{sattler_prior_2017} remains as a reminder that significant improvements need to be made before CNN techniques can be considered competitive when localizing images.

However, CNN-based approaches do possess number of characteristics which qualify them to handle this task well. Namely, CNNs are robust to changes in illumination and occlusion \cite{melekhov_hourglass_2017}, they can operate in close to real time \cite{massiceti_rf_nn_2016} ($\sim 30$ frames per second) and can be trained from labelled data (which can easily be gathered via Structure from Motion (SfM) for any arbitrary scene \cite{schonberger_sfm_2016, schonberger_pixelwise_2016}). CNN based systems also tend to excel in textureless environments where SIFT based methods would typically fail \cite{brachmann_lessmore_2017}. They are also proven to operate well using purely RGB image data --- making them an ideal solution for localizing small, cheap, robotic devices such as drones and unmanned ground vehicles. The major concern of this work is to extend existing pipelines whilst ensuring that the benefits provided by CNNs are \textit{preserved}.

A key observation when considering existing CNN approaches is how position and rotation are treated separately in the loss function. It can be observed that altering a camera's position \textit{or} rotation both affect the image produced, and hence the error in the regressed position and the regressed rotation cannot be decoupled --- each mutually affects the other. In order to optimize a CNN for regressing a camera's pose accurately, a loss term should be used which combines both distinct quantities in an intuitive fashion.

This publication thus offers the following key contributions:
\begin{enumerate}
  \item The formulation of a loss term which considers the error in both the regressed position \textit{and} rotation (Section~\ref{sec:design}). 
  \item Comparison of a CNN trained with and without this loss term on common RGB image localization datasets (Section~\ref{sec:results}).
  \item An indoor image localization dataset (the \textit{Gemini} dataset) with over $3000$ pose-labelled images per-scene (Section~\ref{ssec:datasets}).
\end{enumerate}

%%%%%%%%% RELATED WORK
\section{Related work}

This work builds chiefly on the PoseNet architecture (a camera pose regression network \cite{kendall_posenet_2015}). PoseNet was one of the first CNNs to regress the 6 degrees of freedom in a camera's pose. The network is pretrained on object detection datasets in order to maximize the quality of feature extraction, which occurs in the first stage of the network. It only requires a single RGB image as input, unlike other networks \cite{radwan_vloc_2018, radwan_vloc++_2018}, and operates in real time.  %Additionally, it operates in real time, does not use RANSAC loops, nor does it require solving of the perspective-n-point problem --- unlike the state-of-the-art LessMore network \cite{brachmann_lessmore_2017}.

Notably, PoseNet is able to localize traditionally difficult-to-localize images, specifically those with large textureless areas (where SIFT-based methods fail). PoseNet's end-to-end nature and relatively simple `one-step' training process makes it perfect for the purpose of modification, and in the case of this work, this comes in the form of changing its loss function.

PoseNet has had its loss function augmented in prior works. In \cite{kendall_geometric_2017} it was demonstrated that changing a pose regression network's loss function is sufficient enough to cause an improvement in performance. The network was similarly `upgraded' in \cite{walch_image_2016} using LSTMs to correlate features at the CNN's output. Additional improvements to the network were completed in \cite{kendall_uncertain_2015}, where a Bayesian CNN implementation was used to estimate re-localization accuracy. 

More complex CNN approaches do exist \cite{melekhov_hourglass_2017, melekhov_relative_cnn_2017, purkait_sppnet_2017}. For example, the pipeline outlined in \cite{laskar_camera_2017} uses a CNN to regress the relative poses between a set of images which are \textit{similar} to a query image. These relative pose estimates are coalesced in a fusion algorithm which produces an estimate for the camera pose of the query image.

Depth data has also been incorporated into the inputs of pose regression networks (to improve performance by leveraging multi-modal input information). These RGB-D input pipelines are commonplace in the image localization literature \cite{brachmann_lessmore_2017}, and typically boast higher localization accuracy at the cost of requiring additional sensors, data and computation.

A variety of non-CNN solutions exist, with one of the more notable solutions being the Active Search algorithm \cite{sattler_prior_2017}, which uses SIFT features to inform a matching process. SIFT descriptors are calculated over the query image and are directly compared to a known 3D model's SIFT features. SIFT and other non-CNN learned descriptors have been used to achieve high localization accuracy, but these descriptors tend to be susceptible to changes in the environment, and they often necessitate systems with large amounts of memory and computational power (comparatively to CNNs) \cite{kendall_posenet_2015}. 

% While these prior results motivate our work, we are unable to draw direct comparisons between some of these pipelines and our proposed model. Such pipelines often deal with different input modalities and architectures; the CNN outlined in \cite{kendall_geometric_2017} requires RGB-D or other sensory data in order to estimate scene geometry. 

The primary focus of this work is quantifying the impact of the loss function when training a pose regression CNN. Hence, we do not draw direct comparisons between the proposed model and significantly different pipelines --- such as SIFT-based feature matching algorithms or PoseNet variations with highly modified architectures. Moreover, for the purpose of maximizing the number of available benchmark datasets, we consider pose regressors which handle \textit{purely} RGB query images. In this way, this work deals specifically with CNN solutions to the \textit{monocular}, RGB-only image localization task.

%%%%%%%%% TOPIC
\section{Formulating the proposed loss term}
\label{sec:design}

When trying to accurately regress one's pose based on visual data alone, the error in the two terms which define pose --- position and rotation --- obviously needs to be minimized. If these error terms were entirely minimized, the camera would be in the correct location and would be `looking' in the correct direction. 

Formally, pose regression networks --- such as the default PoseNet --- are trained to regress an estimate $\predpose$ for a camera's true pose $\gtpose$. They do this by calculating the loss after every training iteration, which is formulated as the MSE between the predicted position $\predpos$ and the true position $\gtpos$, plus the MSE between the predicted rotation $\predrot$ and the true rotation $\gtrot$. Note that rotations are encoded as quaternions, since the space of rotations is continuous, and results can be easily normalized to the unit sphere in order to ensure valid rotations. Hyperparameters $\alpha$ and $\beta$ control the balance between positional and rotational error, as illustrated in Equation~\eqref{eq:origlossfn}. In practice, RGB-only pose regression networks reach a maximum localization accuracy when minimizing these error terms independently.

\begin{flalign}
\mathcal{L}_{default} &= \alpha \cdot \| \predpos - \gtpos \| + \beta \cdot \| \predrot - \gtrot \|
\label{eq:origlossfn}
\end{flalign}

\begin{figure}
    \begin{center}
        \fbox{\includegraphics[width=0.95\linewidth]{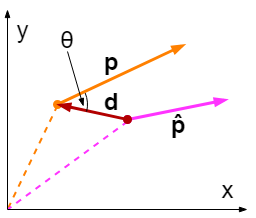}}
        
    \end{center}
    
    \caption{The important quantities required in the calculation of the proposed loss term in 2D. This process naturally extends to 3D. The Euclidean dot product formula is used to calculate a value for $\theta$.}
    \label{fig:termcalc}
\end{figure}

Rather than considering position and rotation as two separate quantities, we consider them together as a line in 3D space: the line travels in a direction defined by the rotation, and must travel through the position vector defined by the position $\gtpos$. We then introduce a `line-of-sight' term which constrains our predictions to lie on this line. The line-of-sight term considers the cosine similarity between the direction of the pose $\gtpose$ and the direction of the difference vector $\deltavec = \gtpos - \predpos$, as per Equation~\eqref{eq:lossterm} and Figure~\ref{fig:termcalc}. This term is only zero when the predicted position lies on the line defined by the ground truth pose, hence constraining the pose regression objective further. In the context of image localization, this ensures that the predicted poses lie on the line-of-sight defined in the ground truth image.

\begin{flalign}
1 - \cos{\theta} = 1 - \frac{\gtpose \cdot \deltavec}{\| \gtpose \| \cdot \| \deltavec \|}
\label{eq:lossterm}
\end{flalign}

We modify the default loss function presented in Equation~\eqref{eq:origlossfn} by adding a weighted contribution of the line-of-sight loss term, producing the proposed loss function in Equation~\eqref{eq:auglossfn}. In practice, the value of $\gamma$ is chosen to roughly reflect the scale of the scene being considered, and is found via a hyperparameter grid search. Note that the line-of-sight term can contribute to the loss through multiplication, higher order terms, etc. but it was determined that weighted addition produced the best performing networks. 

\begin{flalign}
\mathcal{L}_{proposed} &= {L}_{default} + \gamma \cdot (1 - \cos{\theta})
\label{eq:auglossfn}
\end{flalign}

In short, the final loss function used to train the proposed model (Equation~\eqref{eq:auglossfn}) is the result of an exploration in the space of possible loss terms, and the term's design was informed by task specific observations and experimentation.

%Additional experiments were conducted to quantify the effect of the hyperparameter choices on the performance of the network. Of particular interest is the effect of changing the hyperparameter coefficient for the rotational error. Alterations from the default $1.5$ to $1.2$ and from $1.5$ to $1.8$ decreased median rotational accuracy markedly in each case (decreases of $2.3\degree$ on the 7Scenes \textit{Chess} scene). In fact, these changes caused median \textit{positional} accuracy to be increased by $21\%$ and $23\%$ respectively. This suggests that the choice of $1.5$ desirably maximizes the median rotational accuracy, hence why this hyperparameter choice appears in the loss function which we use to augment PoseNet and create what will be herein referred to as our \textit{proposed model}. 

%%%%%%%%% EXPERIMENTS
\section{Experiments}
\label{sec:exps}

Our experiments are naturally centred around testing the performance of the proposed model (defined in Section~\ref{sec:design}). This performance is defined with respect to the following criteria: 

\begin{itemize}
    \item \textbf{Accuracy}: the system should be able to regress a camera's pose with a level of positional and rotational accuracy that is competitive with similar classes of algorithms. Accuracy is reported using per-scene and average median positional and rotational error (See Section~\ref{ssec:acc}).

    \item \textbf{Robustness}: the system should be robust to perceptual aliasing, motion blur and other challenges posed by the considered datasets (See Section~\ref{ssec:rob} and Table~\ref{tab:hardframes}). 

    \item \textbf{Time performance}: evaluation should occur in real-time ($\sim 30$ frames per second), such that the system is suitable in hardware limited real-time applications, or on platforms with RGB-only image sensors, \eg on mobile phones (See Section~\ref{ssec:timeperf}).

    %\item \textbf{Consistency}: performance should be consistent across datasets (See Section~\ref{ssec:perfcons}).
\end{itemize}

%High performance with respect to these metrics is required in the context of robotic navigation pipelines --- a key consideration for this publication.

We compare our proposed model against the default PoseNet and other PoseNet variants. %Another key comparison is the performance of ours and other deep learning pipelines against SIFT based pipelines, or other traditional, feature matching methods.

%%%%%%%%% EXPERIMENTS - DATASETS
\subsection{Datasets}
\label{ssec:datasets}

% Samples, descriptions, origins, usefulness, known challenges

The following datasets are used to benchmark model performance. Each scene's recommended train and test split (see Table~\ref{tab:datasetsizes}) is used throughout the following experiments.

% Dataset size table (single column of paper -> use * in begins)
\begin{table}
    \centering
    \begin{tabular}{ l | c | c c }
        
        \hline
        
                    & Extents   &       &       \\ 
        Scene       & (metres)  & \# Train & \# Test  \\ 
        
        \hline
        
        Chess   & $3\times2\times1$     & 4000  & 2000 \\ 
        Fire    & $2.5\times1\times1$   & 2000  & 2000 \\ 
        Heads   & $2\times0.5\times1$   & 1000  & 1000 \\ 
        Office (7Scenes) & $2.5\times2\times1.5$ & 6000  & 4000 \\ 
        Pumpkin & $2.5\times2\times1$   & 4000  & 2000 \\  
        Red Kitchen & $4\times3\times1.5$   & 7000  & 5000 \\ 
        Stairs  & $2.5\times2\times1.5$ & 2000  & 1000 \\ 
        \hline
        Average & $2.7\times1.9\times1.2$ & 3714  & 2429 \\ 
        
        \hline
        
        Office (University) & $7\times4.5$   & 2196 & 1099 \\ 
        Meeting      & $6.5\times2$   & 1701 & 945  \\ 
        Kitchen      & $6\times7$     & 2076 & 990 \\ % Kitchen1 in the data
        Conference   & $5.5\times7.5$ & 1838 & 949 \\ 
        Coffee Room  & $6\times9$     & 2071 & 959 \\ % Kitchen2 in the data
        \hline
        Average      & $6.2\times6$ & 1976 & 988 \\ 
        
        \hline
        
        % Calc'd great court extents directly from data (96.78 - 1.68)x(78.92 - - 2.26)x(3.34 - - 1.86) using
        % cat dataset_train.txt | grep seq | sort -ki -n | head/tail for i = 2, 3, 4
        Great Court  & $95\times80$   & 1532 & 760  \\ 
        King's College & $140\times40$  & 1220 & 343  \\ 
        Old Hospital & $50\times40$   & 895  & 182  \\ 
        Shop Facade  & $35\times25$   & 231  & 103  \\ 
        St Mary's Church & $80\times60$   & 1487 & 530  \\ 
        Street       & $500\times100$ & 3015 & 2923 \\ 
        \hline
        Average      & $150\times58$ & 1397 & 807  \\ 
        
        \hline
        
        Plain  & $3.3\times2.7\times4.6$   & 2288 & 754 \\ 
        Decor & $3.3\times2.7\times4.6$   & 2288 & 1000 \\ 
        
        \hline
        Average      & $3.3\times2.7\times4.6$ & 2288 & 877  \\ 
        
        \hline
        
    \end{tabular}
    \vspace{-0.3cm}
    \caption{ The size metrics of the \textit{7Scenes}, \textit{University}, \textit{Cambridge Landmarks} and \textit{Gemini} datasets. ($x \times y \times z$) dimensions are given where possible, otherwise ($x \times z$) dimensions are given (where the $y$ axis is the axis perpendicular to the ground). }
    \label{tab:datasetsizes}
    
\end{table}

% Dataset samples
\begin{figure*}[h]
    \centering
    
    % 7Scenes
    \begin{tabularx}{\linewidth}    % Makes it full length
        { *{7}{P{2.1cm}} }
        
        \includegraphics[width=2.1cm]{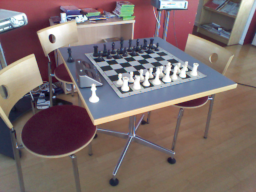} & 
        \includegraphics[width=2.1cm]{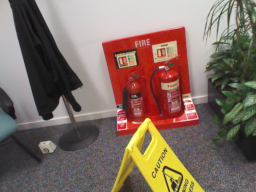} &
        \includegraphics[width=2.1cm]{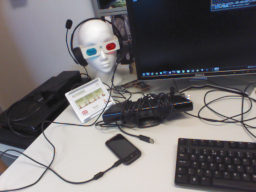} &
        \includegraphics[width=2.1cm]{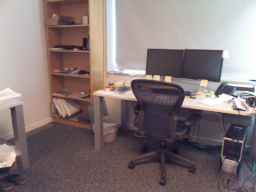} &
        \includegraphics[width=2.1cm]{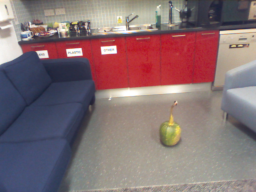} &
        \includegraphics[width=2.1cm]{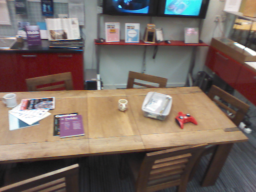} &
        \includegraphics[width=2.1cm]{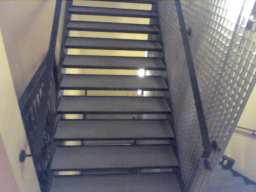} \\
        
        Chess & Fire & Heads & Office & Pumpkin & Red Kitchen & Stairs \\
        
    \end{tabularx}
    \vspace{-0.4cm}
    \caption{ Sample images from each of the 7 scenes in the \textit{7Scenes} dataset. }
    \vspace{0.1cm}
    \label{tab:sample7scenes}
    
    % Cambridge Landmarks
    \begin{tabularx}{\linewidth}    % Makes it full length
        { *{6}{P{2.56cm}} }

        \includegraphics[width=2.56cm]{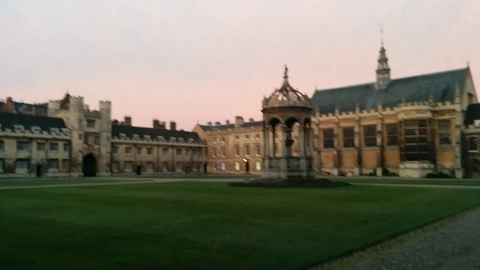} & 
        \includegraphics[width=2.56cm]{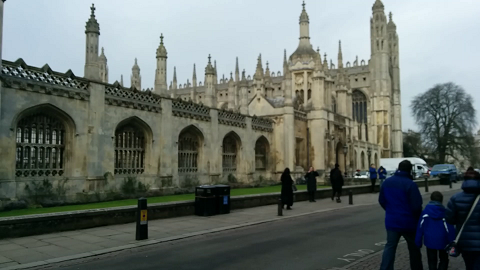} &
        \includegraphics[width=2.56cm]{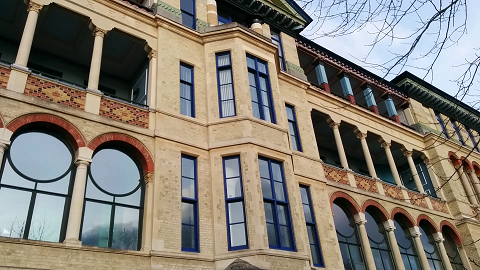} &
        \includegraphics[width=2.56cm]{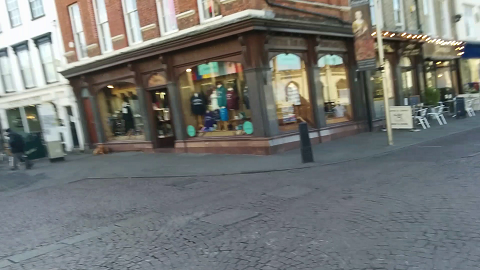} &
        \includegraphics[width=2.56cm]{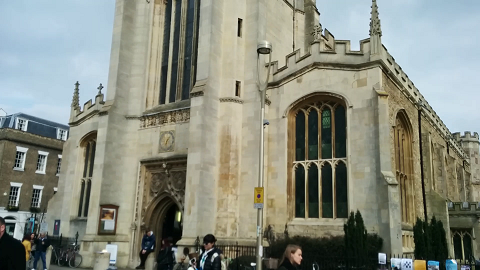} &
        \includegraphics[width=2.56cm]{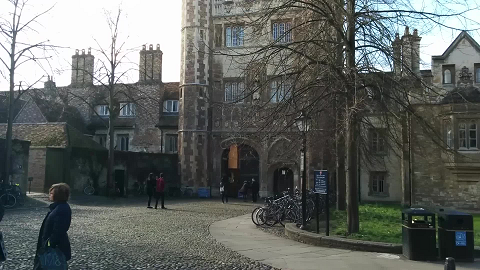} \\
        
        Great Court & Kings College & Old Hospital & Shop Facade & St Mary's Church & Street \\
        
    \end{tabularx}
    \vspace{-0.4cm}
    \caption{ Sample images from each of the 6 scenes in the \textit{Cambridge Landmarks} dataset. }
    \vspace{0.1cm}
    \label{tab:samplecambridge}
    
    % University
    \begin{tabularx}{\linewidth}    % Makes it full length
        { *{5}{P{3.02cm}} }
        
        \includegraphics[width=3.02cm]{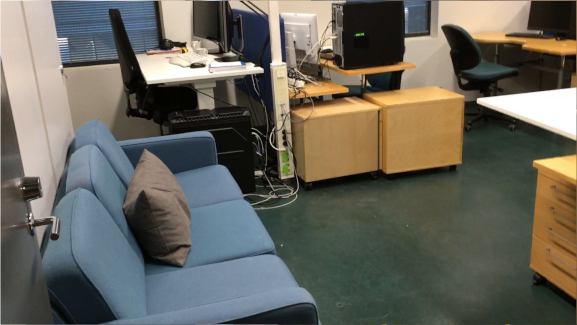} & 
        \includegraphics[width=3.02cm]{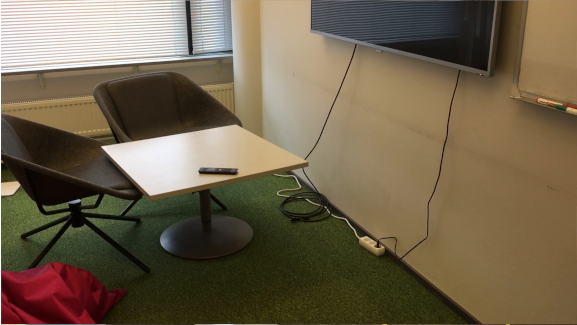} & 
        \includegraphics[width=3.02cm]{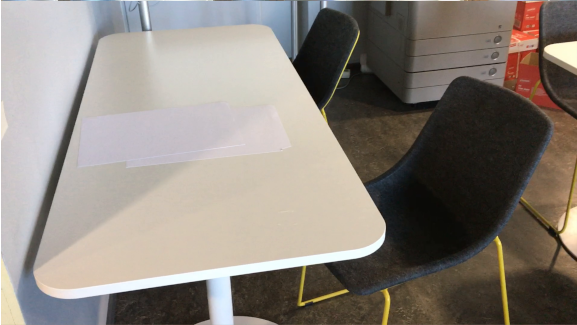} & 
        \includegraphics[width=3.02cm]{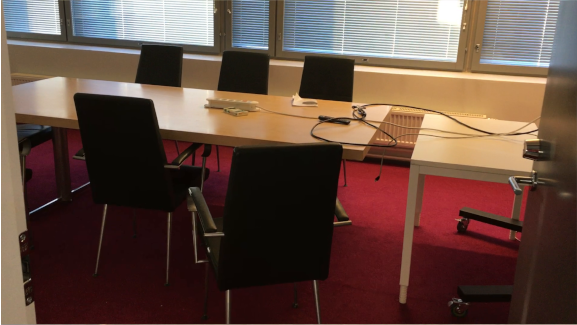} & 
        \includegraphics[width=3.02cm]{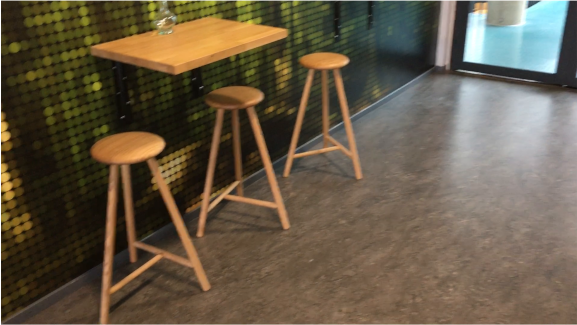} \\
        
        Office & Meeting & Kitchen & Conference & Coffee Room \\
        
    \end{tabularx}
    \vspace{-0.4cm}
    \caption{ Sample images from each of the 5 scenes in the \textit{University} dataset. }
    \vspace{0.1cm}
    \label{tab:sampleuniversity}
    
    % Cambridge Landmarks
    \begin{tabularx}{\linewidth}    % Makes it full length
        { *{6}{P{2.56cm}} }

        \includegraphics[width=2.56cm]{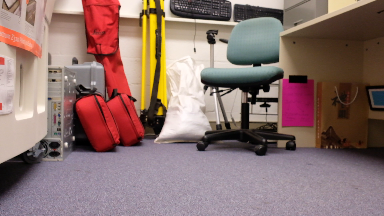} & 
        \includegraphics[width=2.56cm]{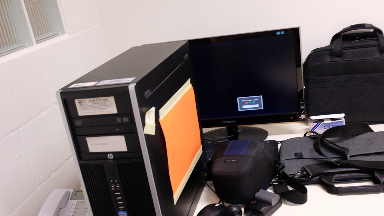} & 
        \includegraphics[width=2.56cm]{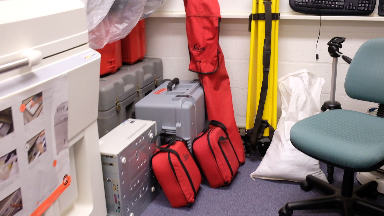} & 
        \includegraphics[width=2.56cm]{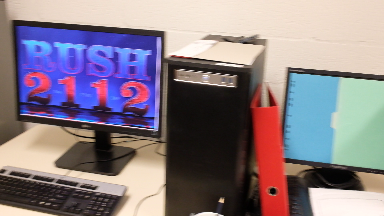} & 
        \includegraphics[width=2.56cm]{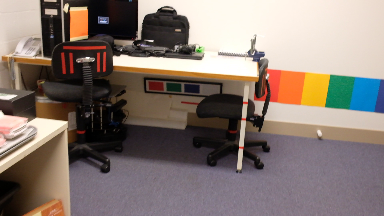} & 
        \includegraphics[width=2.56cm]{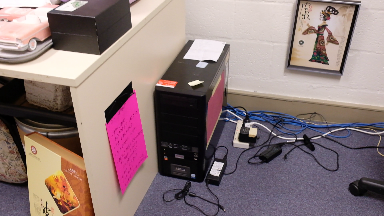} \\
        
    \end{tabularx}
    \vspace{-0.4cm}
    \caption{  Sample images from the 2 scenes in the \textit{Gemini} dataset.}
    \vspace{-0.4cm}
    \label{tab:sampleuwa}
    
\end{figure*}

\textbf{7Scenes} \cite{shotton_7scenes_2013}. $7$ indoor locations in a domestic office context. The dataset features large training and testing sets (in the thousands). The camera paths move continuously while gathering images in distinct sequences. Images include motion blur, featureless spaces and specular reflections (see Figure~\ref{tab:hardframes}), making this a challenging dataset, and one that has been used prolifically in the image localization literature. The ground truths poses are gathered with KinectFusion, and the RGB-D frames each have resolutions of $640\times480$px. 

\textbf{Cambridge Landmarks} \cite{kendall_posenet_2015, kendall_uncertain_2015, kendall_posenet_2015}. $6$ outdoor locations in and around Cambridge, The United Kingdom. The larger spatial extent and restricted dataset size make this a challenging dataset to learn to regress pose from --- methods akin to the one presented in this work typically only deliver positional accuracy in the scale of metres. However, the dataset does provide a common point of comparison, and also includes large expanses of texture-less surfaces. Ground truth poses are generated by a SfM process, so some comparison can be drawn between this dataset and the one created in this work.

\textbf{University} \cite{laskar_camera_2017}. $5$ indoor scenes in a university context. Ground truth poses are gathered using odometry estimates and ``manually generated location constraints in a pose-graph optimization framework'' \cite{laskar_camera_2017}. The dataset, similarly to \textit{7Scenes}, includes challenging frames with high degrees of perceptual aliasing, where multiple frames (with different poses) give rise to similar images \cite{zaval_aliasing_2010}. Although the scenes are registered to a common coordinate system in the \textit{University} dataset and thus a network \textit{could} be trained on the full dataset, the models created in this work are trained and tested \textit{scene-wise} for the purpose of consistency.

\begin{figure}[h]
    \begin{center}
    \begin{tabular}{ *{2}{P{4cm}} }
        
        \includegraphics[width=4cm]{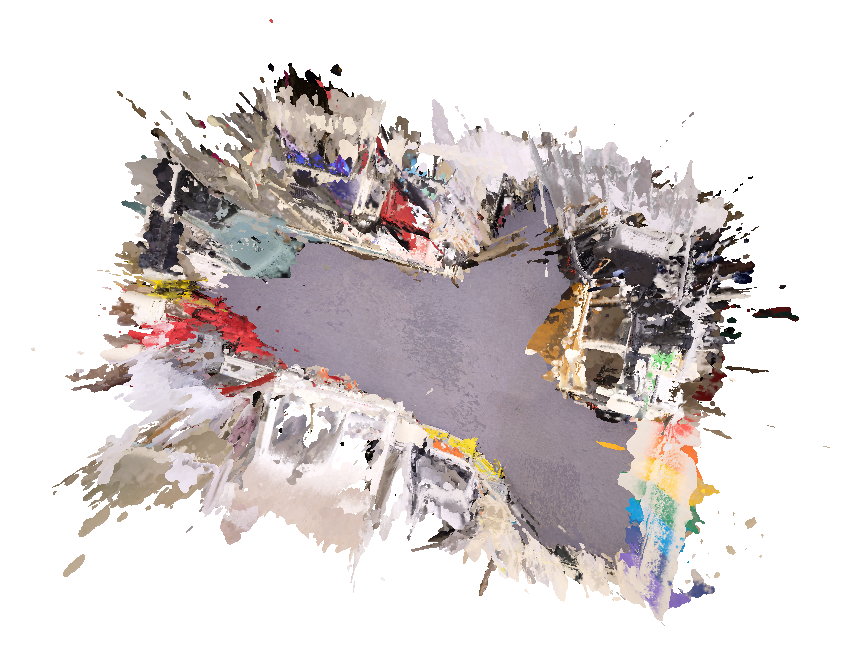} & 
        \includegraphics[width=4cm]{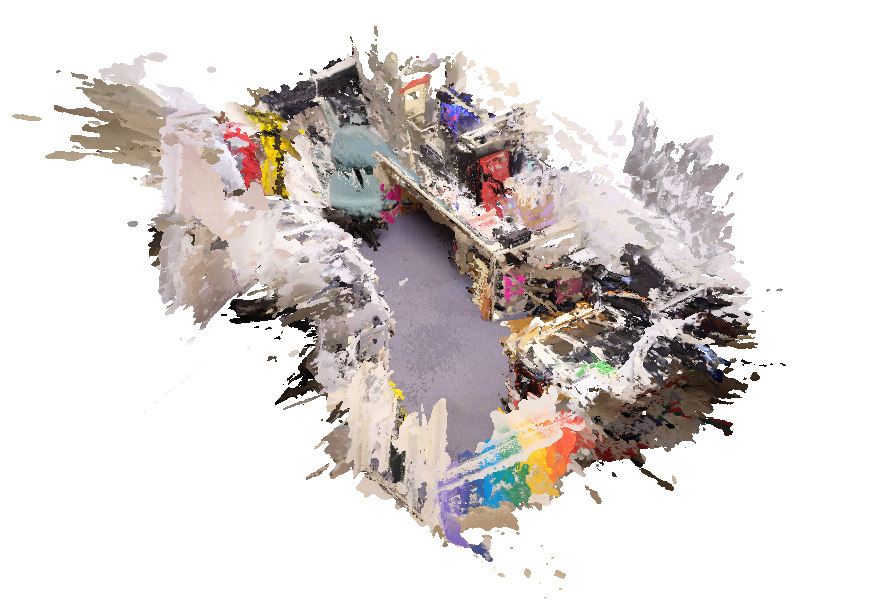} \\
        
        (a) Top down view & (b) Isometric view \\
        
    \end{tabular}
    \end{center}
    \vspace{-0.4cm}
    \caption{ (a) - (b) Varying views of the \textit{Gemini} dataset. }
    \label{fig:uwalab}
    \vspace{-0.4cm}
    
\end{figure}

\textbf{Gemini}\footnote{This dataset has been made available at \url{https://github.com/anon-datasets/gemini}}. $2$ indoor scenes in a university lab context. This dataset was created for the purpose of studying the effect of texture and colour on pose regression networks: both scenes survey the same environment, with one scene including decor (posters, screen-savers, paintings etc.) and the other deliberately \textit{not} including visually rich, textured, and colorful decor. As such the two scenes are labelled \textit{Decor} and \textit{Plain}. A photogrammetry pipeline (COLMAP \cite{schonberger_sfm_2016}) was used to generate the ground truth poses. Images were captured in $15$ separate video sequences using a FujiFilm X-T20 with a 23mm prime autofocus lens (in order to ensure a fixed calibration matrix between sequences). Visualizations of the \textit{with decor} scene are provided in Figure~\ref{fig:uwalab}.

% Particularly difficult test frames from the datasets used

% TODO, some SfM pics and dataset details

%%%%%%%%% EXPERIMENTS - ARCHITECTURE++
\subsection{Architecture and training}
\label{ssec:archtrain}

As stated, we primarily experiment with the PoseNet architecture (using TensorFlow). For the purpose of brevity we redirect the reader to the original publication \cite{kendall_posenet_2015}, as here we only describe crucial elements of the network's design and operation. 

The PoseNet architecture is in itself based on the GoogLeNet architecture \cite{szegedy_googlenet_2014}, a $22$ layer deep network which performs classification and detection. PoseNet extracts GoogLeNet's early feature extracting layers, and replaces the final three softmax classifiers with affine regressors. The network is pretrained using large classification datasets such as \textit{Places} \cite{zhou_places_2014}.

% \cite{kentsommer_tf_2019}
Strictly, the default loss function used is not exactly as defined in Equation~\eqref{eq:origlossfn}. Instead, PoseNet uses the predictions from all three affine regressors (hence there are three predictions for each quantity). We label the $i^{th}$ affine regressor's hyperparameters and predictions using a subscript $i$, as per Equation~\eqref{eq:lossfntrue}. All three affine regressors' predictions are used in the loss function, but each have different hyperparameter weightings: $\alpha_{1}=\alpha_{2}=0.3$, $\alpha_{3}=1$, $\beta_{1}=\beta_{2}=150$ and $\beta_{3}=500$.

\begin{flalign}
\mathcal{L}_{default} &= \alpha_{i} \cdot \| \predpos_{i} - \gtpos \| + \beta_{i} \cdot \| \predrot_{i} - \gtrot \|
\label{eq:lossfntrue}
\end{flalign}

In order to demonstrate the consistency and generalization of the proposed network, we train against all scenes in all datasets using the same experimental setup. For each scene we train PoseNet using the default loss (Equation~\eqref{eq:lossfntrue}) and the proposed loss (Equation~\eqref{eq:auglossfn}) \textit{with} the contribution from all three affine regressors. Each model is trained per-scene over $300,000$ iterations with a batch size of $75$ on a Tesla K40c, which takes $\sim 10$ hours to complete.

%%%%%%%%% RESULTS
\section{Results}
\label{sec:results}

We compare our proposed model to PoseNet and one of its variants --- Bayesian PoseNet \cite{walch_image_2016} --- in Table~\ref{tab:poseacc1}. This is to show the proposed model's performance when compared to other variants of PoseNet with modified loss functions. We then provide results specifically comparing the default PoseNet to our proposed model in Table~\ref{tab:poseacc2}. A discussion of our system's performance regarding the criteria outlined in Section~\ref{sec:exps} follows.

% Table for main pose regression results of different models
\begin{table}
    \centering
    \begin{tabular}{ l | c c c }
        
        \hline
              & Bayesian                                & Default                               & Proposed           \\ 
        Scene & PoseNet \cite{kendall_uncertain_2015}   & PoseNet \cite{kendall_posenet_2015}   & model              \\ 
        
        \hline
        
        Chess               & $0.37$, $7.24$ & $0.32$, $8.12$ & $\bm{0.31}$, $\bm{7.04}$ \\ 
        Fire                & $\bm{0.43}$, $13.7$ & $0.47$, $14.4$ & $0.49$, $\bm{13.3}$ \\ 
        Heads               & $0.31$, $\bm{12.0}$ & $0.29$, $\bm{12.0}$ & $\bm{0.24}$, $15.7$ \\ 
        Office              & $0.48$, $8.04$ & $0.48$, $\bm{7.68}$ & $\bm{0.40}$, $10.0$ \\ 
        Pumpkin             & $0.61$, $\bm{7.07}$ & $\bm{0.47}$, $8.42$ & $0.49$, $9.50$ \\  
        Red Kit.         & $0.58$, $\bm{7.54}$ & $0.58$, $11.3$ & $\bm{0.53}$, $7.98$ \\ 
        Stairs              & $\bm{0.48}$, $\bm{13.1}$ & $0.56$, $15.4$ & $\bm{0.48}$, $14.7$ \\ 
        \hline
        Average             & $0.47$, $\bm{9.81}$ & $0.45$, $11.0$ & $\bm{0.42}$, $11.2$ \\ 
        
        \hline
        
        % Calc'd great court extents directly from data (96.78 - 1.68)x(78.92 - - 2.26)x(3.34 - - 1.86) using
        % cat dataset_train.txt | grep seq | sort -ki -n | head/tail for i = 2, 3, 4
        Street              & ---                    & $\bm{3.67}$, $\bm{6.50}$ & --- \\ 
        King's Col.      & $\bm{1.74}$, $4.06$ & $1.92$, $5.40$ & $2.28$, $\bm{4.05}$ \\ 
        Old Hosp.        & $2.57$, $\bm{5.14}$ & $\bm{2.31}$, $5.38$ & $3.90$, $8.75$ \\ 
        Shop Fac.         & $\bm{1.25}$, $\bm{7.54}$ & $1.46$, $8.08$ & $2.48$, $10.2$ \\ 
        St Mary's       & $\bm{2.11}$, $8.38$ & $2.65$, $8.48$ & $3.02$, $\bm{7.79}$ \\ 
        \hline
        Average$^{1}$       & $\bm{1.92}$, $\bm{6.28}$ & $2.09$, $6.84$ & $2.92$, $7.70$ \\ 
        
        \hline
        
    \end{tabular}
    \vspace{-0.3cm}
    \caption{ The results of various pose regression networks for various image localization datasets. Median positional and rotational error is reported in the form: \textbf{metres, degrees}. The lowest errors are emboldened. Note that our proposed model is competitive in \textit{indoor} datasets with respect to median positional error.
    \newline
    $^{1}$ Average calculated using only the scenes: \textit{King's College}, \textit{Old Hospital}, \textit{Shop Facade} \& \textit{St Mary's Church} as full dataset performance is not available for all pipelines.
    }
    \label{tab:poseacc1}
    \vspace{-0.5cm}
\end{table}

% Table for the university dataset results
\begin{table}
    \begin{center}
    \begin{tabular}{ l | c c  }
        
        \hline
              & Default                               &  Proposed    \\ 
        Scene & PoseNet \cite{kendall_posenet_2015}   &  model \\ 
        
        \hline
        
        Office (University) & $1.05$, $16.2$ & $\bm{0.91}$, $\bm{11.0}$ \\ 
        Meeting             & $1.78$, $10.1$ & $\bm{1.30}$, $\bm{9.58}$ \\ 
        Kitchen             & $\bm{1.19}$, $\bm{12.5}$  & $1.25$, $15.5$ \\ % Kitchen1 in the data
        Conference          & $2.88$, $\bm{13.3}$       & $\bm{2.83}$, $15.8$ \\ 
        Coffee Room         & $1.41$, $14.9$       & $\bm{1.21}$, $\bm{13.3}$ \\ % Kitchen2 in the data
        \hline
        Average             & $1.66$, $13.4$ & $\bm{1.50}$, $\bm{13.0}$ \\ 
        
        \hline
        
        Plain & $1.27$, \bm{$7.87$} & \bm{$1.14$}, $7.90$ \\
        Decor   & $0.15$, $1.17$ & \bm{$0.11$}, \bm{$0.89$} \\
        
        \hline
        
        Average              &   $0.71$, $4.52$ & \bm{$0.63$}, \bm{$4.40$}  \\ 
        
        \hline
        
    \end{tabular}
    \end{center}
    \vspace{-0.5cm}
    \caption{ A study on the direct effects of using our proposed loss function, instead of the default loss function when training PoseNet. Median positional and rotational error is reported in the form: \textbf{metres, degrees}. The lowest errors of each group are emboldened. Note that our contribution majorly outperforms the default PoseNet in both median positional and median rotational error throughout the \textit{University} dataset and the \textit{Gemini} dataset. In the \textit{Gemini} dataset, decreases of $26.7$\% and $24.0$\% in the median positional and rotational error are observed in the \textit{Decor} scene, and an overall increase in accuracy demonstrates the proposed model's robustness to textureless indoor environments (when compared to the default PoseNet).
    }
    %Median translational and rotational error is reported in metres and degrees respectively.
    \label{tab:poseacc2}
    \vspace{-0.5cm}
    
\end{table}

%%%%%%%%%  RESULTS - SPECIFIC
\subsection{Accuracy}
\label{ssec:acc}

It is observed that the proposed model outperforms the default version of PoseNet in approximately half the \textit{7Scenes} scenes --- particularly the \textit{Stairs} scene. In the \textit{Stairs} scene, repetitious structures, \eg staircases, make localization harder, yet the proposed model is robust to such challenges. The network is outperformed in others scenes; namely outdoor datasets with large spatial extents, but in general, performance is improved for the indoor datasets \textit{7Scenes}, \textit{University} and \textit{Gemini}.

A set of cumulative histograms for six of the evaluated scenes are provided in Table~\ref{tab:cumhists}, where we compare the distribution of the positional errors and rotational errors. Median values (provided in Table~\ref{tab:poseacc1} and Table~\ref{tab:poseacc2}) are plotted for reference.

The proposed model's errors are strictly less than the default PoseNet's throughout the majority of the \textit{Chess} and \textit{Coffee Room} distributions. However, the default PoseNet outperforms our proposed model with respect to rotational accuracy in the $10\degree$ - $30\degree$ range in the \textit{Coffee Room} scene.

Note the lesser performance observed from the proposed model on the \textit{King's College} scene; where the positional errors distributions for the two networks are nearly aligned. Moreover, the default PoseNet more accurately regresses rotation in this outdoor scene. See Section~\ref{ssec:rob} and Section~\ref{sec:discfut} for further discussion.

% Cumulative histograms
\begin{table*}[h]
    \centering
    
    % Place in a table
    \begin{tabularx}{\linewidth}    % Makes it full length
        { *{3}{P{5.4cm}} }
        
        \includegraphics[width=5.4cm]{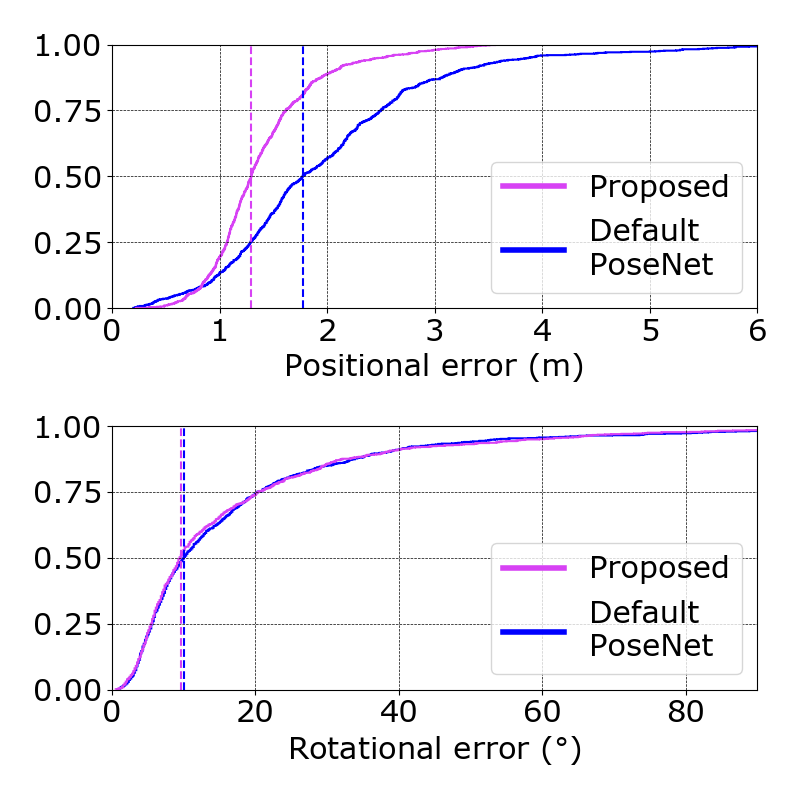} &
        \includegraphics[width=5.4cm]{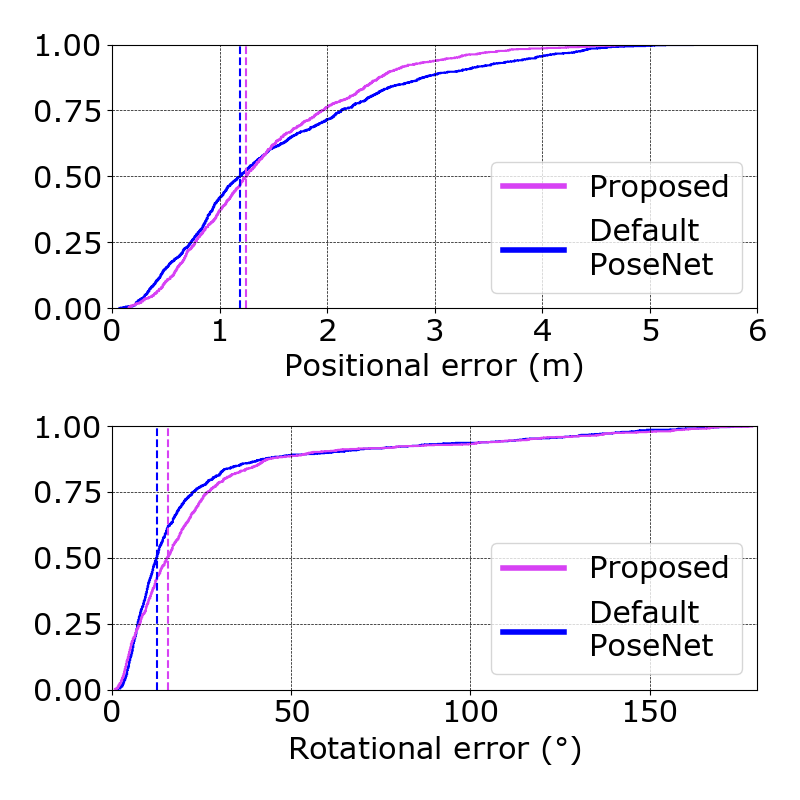} &
        \includegraphics[width=5.4cm]{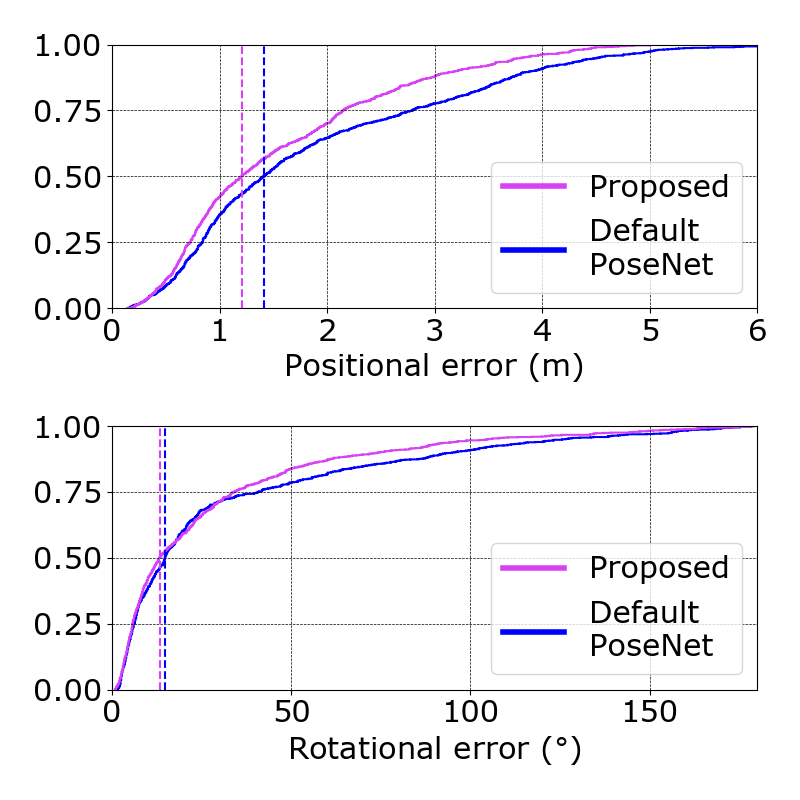} \\
        
        \textit{Meeting} & 
        \textit{Kitchen} & 
        \textit{Coffee Room} \\
        
        \includegraphics[width=5.4cm]{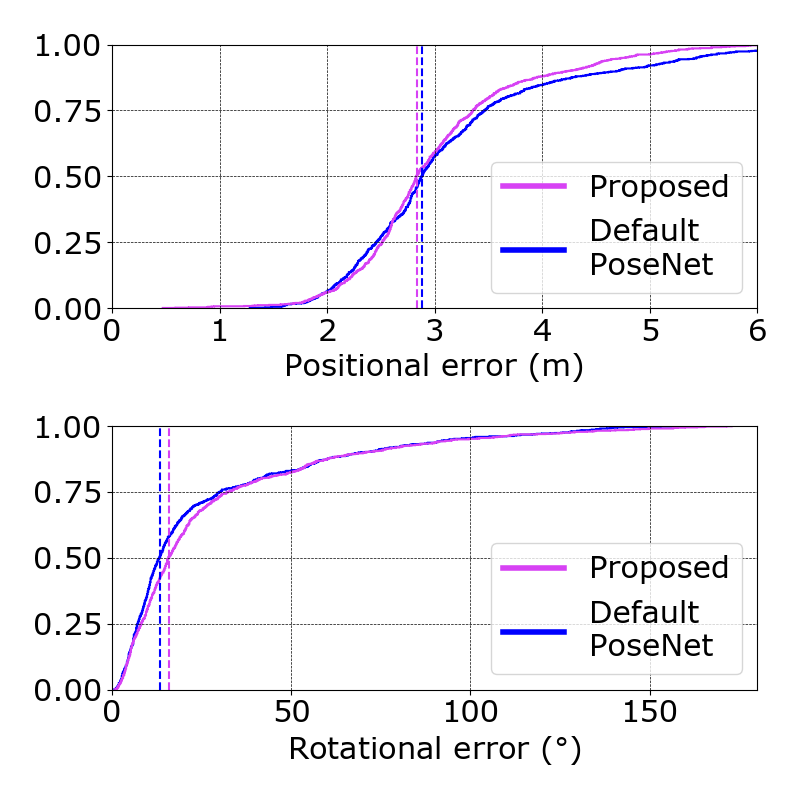} &
        \includegraphics[width=5.4cm]{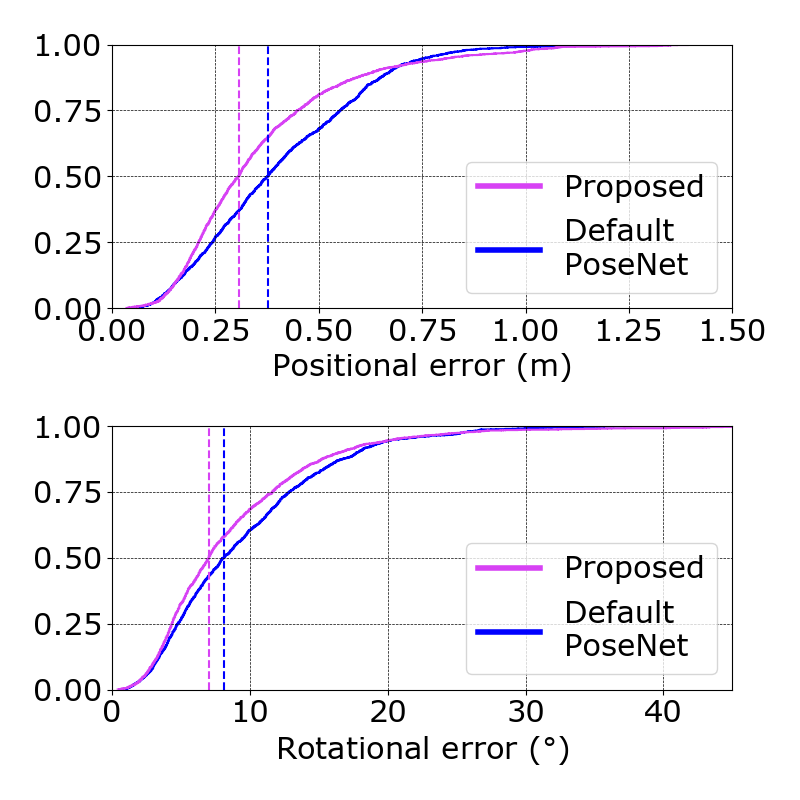} &
        \includegraphics[width=5.4cm]{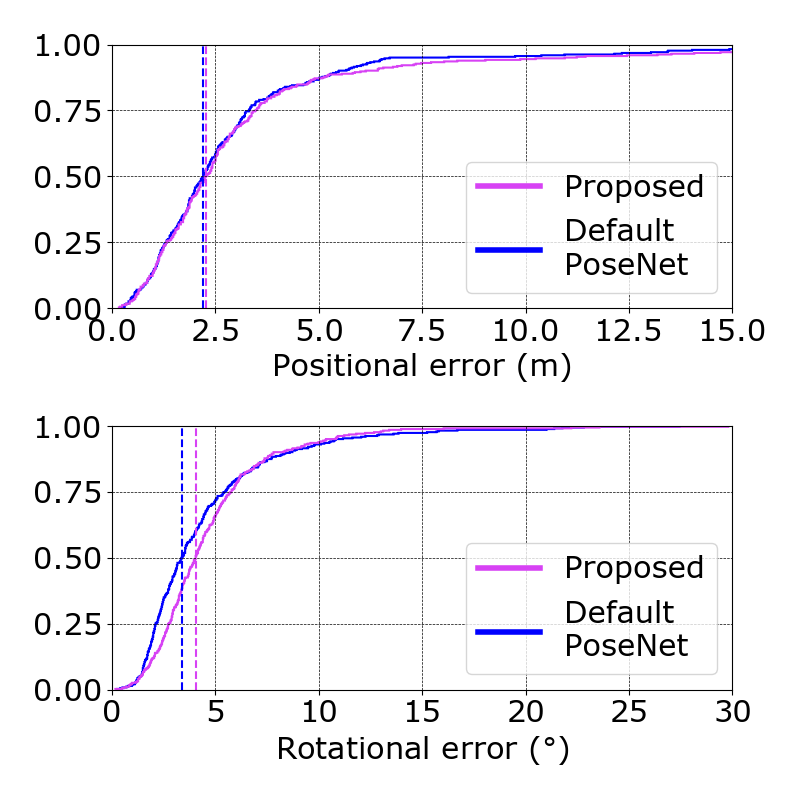} \\
        
        \textit{Conference} & 
        \textit{Chess} & 
        \textit{Kings College} \\
        
    \end{tabularx}
    \caption{ Cumulative histograms of positional and rotational errors, with median values plotted as a dotted line. Note that the proposed model's positional error distribution is strictly less than (shifted to the left of) the default PoseNet's positional error distribution for the indoor scenes (except \textit{Conference}, where performance is comparable). Additionally, the maximum error of the proposed model is lower in the scenes \textit{Meeting, Coffee Room} and \textit{Kitchen}, meaning that our implementation is robust to some of the most difficult frames offered by the \textit{University} dataset. Images best viewed in colour. }
    \label{tab:cumhists}
    
\end{table*}

%%%%%%%%%  RESULTS - SPECIFIC
\subsection{Robustness}
\label{ssec:rob}

\begin{figure}[h]
    \begin{center}
    \begin{tabular}{ *{3}{P{2.4cm}} }
    
        % motion blur --> redkitchen seq 3 000996
        % even better motion blur --> chess seq 5 00040
        % repeated stairs --> stairs seq 1 000119
        % textureless --> pumpkin seq 7 000704/000973
        
        \includegraphics[width=2.4cm]{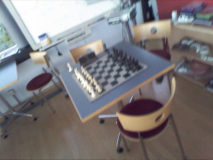} & 
        \includegraphics[width=2.4cm]{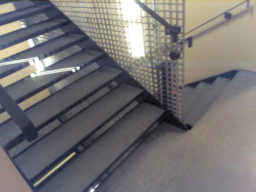} & 
        \includegraphics[width=2.4cm]{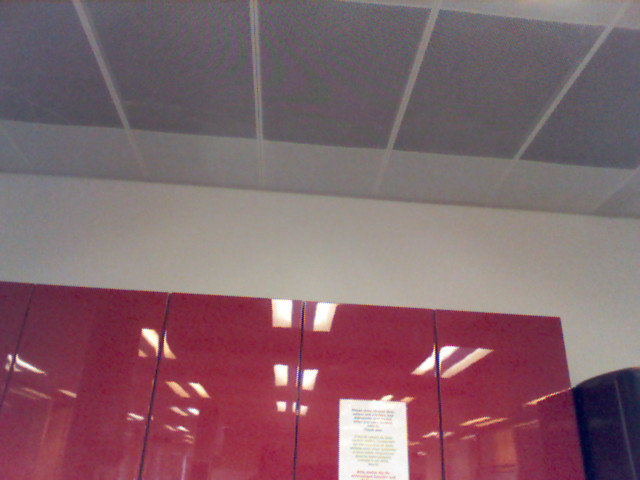} \\
        
        (a) Motion blur & (b) Repeated structures & (c) Textureless \& specular surfaces\\
        
    \end{tabular}
    \end{center}
    \vspace{-0.8cm}
    \caption{ (a) - (c) Images from the \textit{7Scenes} dataset where accurately regressing pose is challenging. }
    \label{tab:hardframes}
    
\end{figure}

The robustness of our system to challenging test frames --- that is, images with motion blur, repeated structures or demonstrating perceptual aliasing \cite{li_fullframe_2018} --- can be determined via the cumulative histograms in Table~\ref{tab:cumhists}. For the purpose of visualization, some difficult testing images from the \textit{7Scenes} dataset are displayed in Figure~\ref{tab:hardframes}. 

The hardest frames in the test set by definition produce the greatest errors. Consider the positional error for the \textit{Meeting} scene: our proposed model reaches a value of $1.0$ on the y-axis before the default PoseNet does, meaning that the \textit{hardest} frames in the test set have their position regressed more accurately. This analysis extends to each of the cumulative histograms in Table~\ref{tab:cumhists}, thus confirming our proposed loss function's robustness to difficult test scenarios, as the frames of greatest error consistently have less than or comparable errors when compared to the default PoseNet. 

%Our model outperforms the default PoseNet across the \textit{Gemini} dataset, illustrating its robustness to textureless indoor environments, which the \textit{Plain} scene features extensively. 

% Particularly difficult test frames from the datasets used

Moreover, the proposed model significantly exceeds the default PoseNet's performance throughout the \textit{Gemini} dataset. The performance gap in the \textit{Plain} scene proves that our model is more robust to textureless spaces than the default PoseNet.

%%%%%%%%%  RESULTS - SPECIFIC
\subsection{Efficiency}
\label{ssec:timeperf}

\textbf{Training time}. The duration of the training stage compared between our implementation and default PoseNet is by design, very similar, and highly competitive when compared to the other systems analyzed in Table~\ref{tab:poseacc1}. This is due to the relatively inexpensive computing cost of introducing a simple line-of-sight loss term into the network's overall loss function. The average training time for default PoseNet and for our augmented PoseNet over the University dataset is $10:21:31$ and $10:23:33$ respectively (HH:MM:SS), where both tests are ran on the same hardware. 

\textbf{Testing time}. The network operation during the test time is naturally not affected by the loss function augmentation. The time performance when testing is similar to that of the default PoseNet and in general is competitive amongst camera localization pipelines (especially feature based matching techniques). We observe a total elapsed time of $16.04$ seconds when evaluating the entire \textit{Coffee Room} scene testing set, whereas it takes $16.03$ seconds using the default PoseNet. In other words, both systems take $\sim16.8$ ms to complete a single inference on our hardware.

\textbf{Memory cost}. Memory cost in general for CNNs is low --- only the weights for the trained layers and the input image need to be loaded into memory. When compared to feature matching techniques, which need to store feature vectors for all instances in the test set, or SIFT-based matching methods with large memory and computational overheads, CNN approaches are in general quite desirable --- especially in resource constrained environments. Both the proposed model and the default PoseNet take $8015$MiB and $10947$MiB to train and test respectively (as reported by \textit{nvidia-smi}). For interest, the network weights for the proposed model's TensorFlow implementation total only $200$MB.

%%%%%%%%% DISCUSSION & FUTURE WORK 
\section{Discussion and future work}
\label{sec:discfut}

Experimental results confirm that the proposed loss term has a positive impact on robustness and accuracy, whilst maintaining speed, memory usage, and robustness (to textureless spaces and so forth). %The proposed model could still be improved with regards to its relative accuracy compared to other, more advanced/complex pipelines \cite{kendall_geometric_2017, walch_image_2016}, but such pipelines demonstrate that an increase in accuracy comes at the expense of time performance. In summary, our model is highly performant in the space of RGB-only, monocular, pose regression networks.

% Statements like this suggest that our work is not mature:
%Whilst the augmented network outlined in this work exhibits comparatively high performance in some areas, it could still be improved in future work.

%For example, the proposed model \textit{is} outperformed by competitors which use more expensive reprojection methods \cite{kendall_geometric_2017}, but these procedures infringe on the default PoseNet's ease-of-training. 

The network \textit{is} outperformed by the SIFT-based image localization algorithm `Active Search' \cite{sattler_prior_2017}, indicating that there is still some work required until the gap between SIFT-based algorithms and CNNs is closed (in the context of RGB-only image localization). However, SIFT localization operates on a much longer timescale, and can be highly computationally expensive depending on the dataset and pipeline being used \cite{wu_linear_sfm_2013}.

%In a similar vein, the proposed model is outperformed by PoseNet augmented with LSTMs \cite{walch_image_2016} --- a more computationally expensive approach --- though it should be noted that our approaches are not necessarily mutually exclusive, and LSTMs with augmented loss functions is perhaps an avenue for future research.

Ultimately, the loss function described in this work illustrates that intuitive loss terms, designed with respect to a specific task (in this case image localization) can positively impact the performance of deep networks.

%As an aside, we note a discrepancy in the consistency of our model between datasets with differing spatial extents. It is suspected that further hyperparameter tuning is required in order to allow the $\mathcal{L}_{geo}$ term (Equation~\eqref{eq:lossterm}) to balance better with the default positional and rotational loss terms (Equation~\eqref{eq:origlossfn}). This would suggest that the loss term introduces a dependency on the average scene geometry, but more investigation is required in order to provide a meaningful discussion of these results. We leave this investigation up to future work. % The focus for this model however is on indoor datasets, where data is abundant, easily gathered and environments are more structured (ideal for robotics applications and other CV tasks).

Possible avenues for future work include extending this loss function design methodology to other CV tasks, in order to achieve higher performance, or to consider RGB-D pipelines. An investigation on the effect that such loss terms have on the convergence rate, and upper performance limit of NNs could also be explored.

%%%%%%%%% CONCLUSION
\section{Conclusion}

In summary, the effect of adding a line-of-sight loss term to an existing pose regression network is investigated. The performance of the proposed model is compared to other similar models across common image localization benchmarks and the newly introduced \textit{Gemini} dataset. Improvements to performance in the image localization task are observed, without any drastic increase in evaluation speed or training time. Particularly, the median positional accuracy is --- on average --- increased for indoor datasets when compared to a version of the model without the suggested loss term. 

This work suggests that means squared error between the ground truth and the regressed predictions --- although often used as a measure of loss for many Neural Networks --- can be improved upon. Specifically, loss functions designed with the network's task in mind may yield better performing models. For pose regression networks, the distinct and coupled nature of positional and rotational quantities needs to be considered when designing a network's loss function.

%%%%%%%%% REFERENCES
% No page limit
% Following line references all publications for the purpose of editing
% \nocite{*}

{\small
\bibliographystyle{ieee}
\bibliography{egbib}
}

\end{document}